\documentclass[runningheads]{llncs}
\IfFileExists{lmodern.sty}{\usepackage[T1]{fontenc}\usepackage{lmodern}}{}
\usepackage{graphicx}
\usepackage{rotating}
\usepackage{amsmath,amssymb}
\usepackage{booktabs}
\usepackage{tabularx}
\newcolumntype{Y}{>{\raggedright\arraybackslash}X}
\usepackage{multirow}
\usepackage{xcolor}
\usepackage{tikz}
\usetikzlibrary{arrows.meta,positioning,shapes.geometric,fit,calc}
\usepackage[hidelinks]{hyperref}
\usepackage{xspace}
\IfFileExists{lmodern.sty}{\usepackage{microtype}}{\usepackage[expansion=false]{microtype}}
\newcommand{\nLabOrig}{46}
\newcommand{\nLabRet}{25}
\newcommand{\minPos}{10}

\newcommand{\nTest}{640}

\newcommand{\leakExact}{0}
\newcommand{\errTest}{35.8}

\newcommand{\odirMapped}{ARMD, BRVO, CRVO, CSR, DN, DR, ERM, LS, MHL, MYA, ODE, TSLN}
\newcommand{\nIdrid}{81}
\newcommand{\nBoot}{1000}
\newcommand{\aucRf}{0.925}
\newcommand{\scrRf}{0.974}
\newcommand{\foneRf}{0.477}
\newcommand{\aucRfLo}{0.909}
\newcommand{\aucRfHi}{0.941}
\newcommand{\aucCnx}{0.952}
\newcommand{\scrCnx}{0.981}
\newcommand{\foneCnx}{0.508}
\newcommand{\aucCnxLo}{0.939}
\newcommand{\aucCnxHi}{0.963}
\newcommand{\aucSwin}{0.949}
\newcommand{\scrSwin}{0.983}
\newcommand{\foneSwin}{0.513}
\newcommand{\aucSwinLo}{0.935}
\newcommand{\aucSwinHi}{0.961}
\newcommand{\aucRes}{0.897}
\newcommand{\scrRes}{0.964}
\newcommand{\foneRes}{0.408}
\newcommand{\aucResLo}{0.870}
\newcommand{\aucResHi}{0.920}
\newcommand{\aucEff}{0.874}
\newcommand{\scrEff}{0.971}
\newcommand{\foneEff}{0.446}
\newcommand{\aucEffLo}{0.841}
\newcommand{\aucEffHi}{0.901}
\newcommand{\aucRfCons}{0.896}
\newcommand{\scrRfCons}{0.958}
\newcommand{\foneRfCons}{0.432}
\newcommand{\aucRfConsLo}{0.869}
\newcommand{\aucRfConsHi}{0.919}
\newcommand{\aucRfSeedMean}{0.922}
\newcommand{\aucRfSeedSd}{0.007}
\newcommand{\eceRaw}{0.123}
\newcommand{\eceCal}{0.126}

\newcommand{\errAucEnt}{0.704}
\newcommand{\errAucMI}{0.633}

\newcommand{\aucAz}{0.685}
\newcommand{\edrAz}{29.7}
\newcommand{\aucAone}{0.747}
\newcommand{\edrAone}{37.1}
\newcommand{\aucAtwo}{0.751}
\newcommand{\edrAtwo}{35.4}
\newcommand{\aucAthree}{0.757}
\newcommand{\edrAthree}{33.2}
\newcommand{\aucAfour}{0.766}
\newcommand{\edrAfour}{38.4}
\newcommand{\dVerif}{0.004}
\newcommand{\dVerifLo}{-0.027}
\newcommand{\dVerifHi}{0.033}

\newcommand{\dQual}{0.006}

\newcommand{\dXai}{0.009}

\newcommand{\nErrTest}{229}
\newcommand{\capBoneTen}{33}

\newcommand{\srBoneTen}{34.0}

\newcommand{\capBtwoTen}{35}

\newcommand{\srBtwoTen}{33.7}

\newcommand{\capBfourTen}{43}

\newcommand{\srBfourTen}{32.3}

\newcommand{\capBfiveTen}{48}

\newcommand{\srBfiveTen}{31.4}

\newcommand{\capBsixTen}{48}

\newcommand{\srBsixTen}{31.4}

\newcommand{\capBoneTwenty}{68}

\newcommand{\srBoneTwenty}{31.4}

\newcommand{\capBtwoTwenty}{70}

\newcommand{\srBtwoTwenty}{31.1}

\newcommand{\capBfourTwenty}{76}

\newcommand{\srBfourTwenty}{29.9}

\newcommand{\capBfiveTwenty}{82}

\newcommand{\srBfiveTwenty}{28.7}

\newcommand{\capBsixTwenty}{88}

\newcommand{\srBsixTwenty}{27.5}

\newcommand{\capBoneThirty}{99}

\newcommand{\srBoneThirty}{29.0}

\newcommand{\capBtwoThirty}{103}

\newcommand{\srBtwoThirty}{28.1}

\newcommand{\capBfourThirty}{115}

\newcommand{\srBfourThirty}{25.4}

\newcommand{\capBfiveThirty}{119}

\newcommand{\srBfiveThirty}{24.6}

\newcommand{\capBsixThirty}{122}

\newcommand{\srBsixThirty}{23.9}

\newcommand{\srBase}{35.8}
\newcommand{\dEdrTen}{6.6}
\newcommand{\dEdrTenLo}{1.7}
\newcommand{\dEdrTenHi}{10.4}

\newcommand{\nBudgets}{6}
\newcommand{\nCiBudgets}{4}

\newcommand{\minHolmTwo}{=0.096}

\newcommand{\pEdrFive}{<0.001}
\newcommand{\pMcFive}{=0.024}
\newcommand{\capBoneFive}{12}

\newcommand{\capBfiveFive}{26}
\newcommand{\covBoneTen}{50.2}
\newcommand{\muBoneTen}{11.1}

\newcommand{\covBtwoTen}{59.1}
\newcommand{\muBtwoTen}{14.2}

\newcommand{\covBfourTen}{52.5}
\newcommand{\muBfourTen}{9.5}

\newcommand{\covBfiveTen}{46.4}
\newcommand{\muBfiveTen}{8.0}

\newcommand{\covBoneFive}{33.1}
\newcommand{\muBoneFive}{5.3}

\newcommand{\covBtwoFive}{38.4}
\newcommand{\muBtwoFive}{7.8}

\newcommand{\covBfourFive}{37.2}
\newcommand{\muBfourFive}{5.0}

\newcommand{\covBfiveFive}{37.7}
\newcommand{\muBfiveFive}{5.5}

\newcommand{\covBoneTwo}{17.5}
\newcommand{\muBoneTwo}{2.5}

\newcommand{\covBtwoTwo}{20.9}
\newcommand{\muBtwoTwo}{2.3}

\newcommand{\covBfourTwo}{22.2}
\newcommand{\muBfourTwo}{2.2}

\newcommand{\covBfiveTwo}{20.0}
\newcommand{\muBfiveTwo}{2.2}

\newcommand{\lttAlpha}{20.0}
\newcommand{\lttCov}{44.8}

\newcommand{\crcMean}{4.9}
\newcommand{\crcQnine}{6.9}

\newcommand{\errblurThree}{72.0}

\newcommand{\riskblurThree}{0.125}

\newcommand{\mucontrastThree}{28.0}

\newcommand{\riskClean}{0.357}
\newcommand{\muNoQmax}{53.1}
\newcommand{\muMax}{28.0}

\newcommand{\scrAucR}{0.974}

\newcommand{\miR}{0.0292}
\newcommand{\riskR}{0.357}

\newcommand{\riskAucR}{0.766}

\newcommand{\scrAucO}{0.663}

\newcommand{\miO}{0.0271}
\newcommand{\riskO}{0.489}
\newcommand{\refO}{87.4}
\newcommand{\muO}{6.8}

\newcommand{\riskAucO}{0.558}

\newcommand{\cdRisk}{0.284}
\newcommand{\pShiftRisk}{<0.001}
\newcommand{\ptCam}{0.864}
\newcommand{\enrCam}{2.24}

\newcommand{\ptIg}{0.815}
\newcommand{\enrIg}{1.71}

\newcommand{\ptRoll}{0.123}
\newcommand{\enrRoll}{0.98}

\newcommand{\stabRoll}{0.95}

\newcommand{\msFixed}{345}
\newcommand{\msAgent}{246}

\newcommand{\msMc}{199}
\newcommand{\msDet}{20}
\newcommand{\msVer}{19}
\newcommand{\msXai}{81}
\newcommand{\msQual}{27}

\newcommand{\vcRwVc}{69}
\newcommand{\vcBoth}{160}

\newcommand{\verAuc}{0.691}

\newcommand{\dUnc}{0.061}
\newcommand{\dUncLo}{0.025}
\newcommand{\dUncHi}{0.096}

\newcommand{\cmpWauc}{0.896}
\newcommand{\cmpWerr}{43.8}
\newcommand{\cmpWAz}{0.614}
\newcommand{\cmpWAone}{0.790}
\newcommand{\cmpWAtwo}{0.848}

\newcommand{\cmpWdVer}{0.058}
\newcommand{\cmpWdVerLo}{0.027}
\newcommand{\cmpWdVerHi}{0.089}
\newcommand{\cmpWpVer}{<0.001}

\newcommand{\cmpWcapBoneTwenty}{88}

\newcommand{\cmpWcapBfiveTwenty}{119}
\newcommand{\cmpWdOneTwenty}{11.1}
\newcommand{\cmpWpOneTwenty}{<0.001}
\newcommand{\cmpWdTwoTwenty}{12.9}
\newcommand{\cmpWpTwoTwenty}{<0.001}

\newcommand{\cmpWnSigOne}{4}
\newcommand{\cmpWnB}{6}
\newcommand{\cmpWnSigTwo}{5}
\newcommand{\cmpWnErr}{280}
\newcommand{\cmpWcovBone}{10.3}

\newcommand{\cmpWcovBtwo}{28.0}

\newcommand{\cmpWcovBfive}{37.0}
\newcommand{\cmpWmuBfive}{5.6}
\newcommand{\cmpWmuMax}{7.7}
\newcommand{\cmpWmuNoQ}{60.2}
\newcommand{\cmpWver}{24.8}
\newcommand{\cmpWsave}{30}
\newcommand{\cmpSauc}{0.925}
\newcommand{\cmpSerr}{35.8}
\newcommand{\cmpSAz}{0.685}
\newcommand{\cmpSAone}{0.747}
\newcommand{\cmpSAtwo}{0.751}

\newcommand{\cmpSdVer}{0.004}
\newcommand{\cmpSdVerLo}{-0.027}
\newcommand{\cmpSdVerHi}{0.033}

\newcommand{\cmpSdOneTwenty}{6.1}
\newcommand{\cmpSpOneTwenty}{0.150}
\newcommand{\cmpSdTwoTwenty}{5.2}
\newcommand{\cmpSpTwoTwenty}{0.200}

\newcommand{\cmpSnSigOne}{1}
\newcommand{\cmpSnB}{6}
\newcommand{\cmpSnSigTwo}{0}
\newcommand{\cmpSnErr}{229}
\newcommand{\cmpScovBone}{33.1}

\newcommand{\cmpScovBtwo}{38.4}

\newcommand{\cmpScovBfive}{37.7}

\newcommand{\cmpSmuMax}{28.0}
\newcommand{\cmpSmuNoQ}{53.1}
\newcommand{\cmpSver}{7.7}
\newcommand{\cmpSsave}{29}

\begin{document}
\title{RetinaGuard: Risk-Controlled Agentic Triage for Multi-Disease Retinal Screening}
\titlerunning{RetinaGuard: Risk-Controlled Agentic Retinal Triage}
\author{Sohaib Sulman\inst{1}\thanks{Corresponding author.} \and
Muhammad Atif Saeed\inst{1} \and
Akhtar Jamil\inst{2} \and
Alaa Ali Hameed\inst{3} \and
Roland Aydin\inst{4,5}}
\authorrunning{S. Sulman et al.}
\institute{Department of Data Science and Artificial Intelligence, National University of Computer and Emerging Sciences,
Islamabad, Pakistan\\
\email{i248068@isb.nu.edu.pk, atif.saeed@isb.nu.edu.pk}
\and
Department of Computer Science, National University of Computer and Emerging Sciences, Islamabad, Pakistan\\
\email{akhtar.jamil@nu.edu.pk}
\and
Department of Computer Engineering, Istinye University, Istanbul, Turkey\\
\email{alaa.hameed@istinye.edu.tr}
\and
Department of Materials Science, Saarland University, Saarbr\"ucken, Germany\\
\email{roland.aydin@uni-saarland.de}
\and
German Research Center for Artificial Intelligence (DFKI), Saarbr\"ucken, Germany}
\maketitle

\begin{abstract}
Retinal screening programmes need to know which automated decisions can be accepted without review. This paper presents RetinaGuard,
an agentic triage framework in which quality, diagnostic, uncertainty, verification and explanation agents gather evidence for each
fundus photograph, a learned failure predictor estimates the probability that the decision is unsafe, auxiliary tools are invoked only
when their expected value of information exceeds their cost, and the acceptance threshold is certified by conformal risk control. On the
RFMiD test split, RetinaGuard was compared with confidence and calibrated-uncertainty referral at matched referral budgets and equal
certified safety, using two RETFound primary models of different strength. With the weaker model (macro-AUROC \cmpWauc), verifier
disagreement added significant information, RetinaGuard captured \cmpWcapBfiveTwenty{} of \cmpWnErr{} unsafe decisions at a 20\% budget
against \cmpWcapBoneTwenty{} for lowest-confidence referral, and at a nominal 5\% unsafe level it accepted \cmpWcovBfive\% of cases
against \cmpWcovBone\%. With the stronger model (\cmpSauc) the verifier added nothing beyond uncertainty and the advantage shrank to a
few percentage points. In both settings its coverage at the nominal 5\% level stayed near \cmpScovBfive\%, gating cut computation by about 30\%, and a
quality agent reduced unsafe acceptances under most image corruptions. The benefit of agentic evidence gathering thus depends on how
much the primary model's own uncertainty already reveals about its errors.
\end{abstract}

\section{Introduction}\label{sec:intro}
Deep learning systems detect diabetic retinopathy from colour fundus photographs at specialist level~\cite{gulshan2016dr}, an
autonomous diagnostic system has been authorised by the US Food and Drug Administration for use in primary care after a pivotal
trial~\cite{abramoff2018pivotal}, and retinal foundation models such as RETFound~\cite{zhou2023retfound} promise transfer to many
diseases from a single pretrained encoder. Screening programmes, however, need more than average accuracy. For every image they must
decide whether the automated decision can be trusted or whether the case should be referred to an ophthalmologist. A multi-disease
screen is safe only if the referred cases contain most of the errors, if the error rate among accepted cases is controlled, and if
this control survives a change of camera or a degraded photograph.

The standard approach ranks cases by confidence or uncertainty and refers the least certain ones~\cite{leibig2017uncertainty}. Recent
agentic clinical systems extend this idea with tools and multi-step reasoning~\cite{kim2024mdagents,fallahpour2025medrax}, but they are
rarely compared with confidence-based referral at the same referral cost, and their decisions carry no statistical guarantee. Adding
tools around a classifier is useful only if the evidence they produce changes the accept or refer decision for the better, and
whether it does is likely to depend on how informative the classifier's own uncertainty already is.

This paper studies that question for multi-disease retinal screening with RetinaGuard, a risk-controlled and cost-aware agentic
triage framework. The main contributions are as follows.
\begin{itemize}
\item An evidence-driven triage agent that combines image quality, calibrated and Monte Carlo uncertainty, disagreement with an
independent verifier and explanation features into a learned probability that an automated decision is unsafe.
\item A value-of-information criterion that invokes the verifier and explanation tools only when their expected effect on the
decision exceeds their measured computational cost.
\item Conformal risk control of the acceptance threshold, which bounds the rate of unsafe automated acceptances for the adaptive
policy and allows all methods to be compared at equal certified safety.
\item A controlled evaluation at matched referral budgets with two primary models of different strength, showing that verification
improves triage substantially for the weaker model and not for the stronger one, while certified coverage, computational savings and
protection against degraded images persist in both settings.
\end{itemize}

The remainder of the paper is organised as follows. Section~\ref{sec:related} reviews related work. Section~\ref{sec:method}
describes the datasets and the proposed framework. Section~\ref{sec:results} presents the experimental setup and results.
Section~\ref{sec:discussion} discusses the findings and limitations, and Section~\ref{sec:conclusion} concludes the paper.

\section{Related Work}\label{sec:related}
Deep learning systems for referable diabetic retinopathy reached specialist-level sensitivity on large retrospective
cohorts~\cite{gulshan2016dr}, and autonomous deployment in primary care was later supported by a prospective pivotal
trial~\cite{abramoff2018pivotal}. Public multi-disease datasets such as RFMiD~\cite{pachade2021rfmid} and ODIR-5K~\cite{li2021odir}
broadened the task to dozens of conditions, and the submissions to the RFMiD challenge showed that the best methods combine
preprocessing, augmentation, pretrained models and ensembling~\cite{pachade2025rfmidchallenge}. Retinal foundation models changed the
starting point of such systems: RETFound was pretrained with masked autoencoding on 1.6 million retinal images and transferred to
ocular and systemic disease detection~\cite{zhou2023retfound}. A recent systematic review reports strong external validation for several
ophthalmic foundation models but no routine clinical deployment~\cite{jin2026fmreview}, and a natural-image foundation model has
outperformed RETFound for ocular disease detection~\cite{hou2026dinov2}. These studies report discrimination at a fixed operating point
and do not decide, case by case, whether an automated decision may stand.

Uncertainty estimation is the usual basis for that decision. Monte Carlo dropout~\cite{gal2016dropout} and deep
ensembles~\cite{lakshminarayanan2017ensembles} approximate predictive uncertainty, temperature scaling corrects
overconfidence~\cite{guo2017calibration}, and the maximum softmax probability remains a strong failure
detector~\cite{hendrycks2017baseline}. Leibig et al.~\cite{leibig2017uncertainty} showed that referring uncertain diabetic retinopathy
predictions improves accuracy on the retained cases. Learning-to-defer methods go further and learn when to pass a case to a clinician;
CoDoC improved the combined accuracy of clinicians and AI in breast cancer and tuberculosis screening~\cite{dvijotham2023codoc}. Most of
this work ranks cases by a single score or learns a deferral rule, but does not ask which additional evidence is worth acquiring
before deciding, nor whether richer evidence improves on uncertainty at a matched referral rate.

Conformal prediction provides distribution-free guarantees for exchangeable data~\cite{angelopoulos2023gentle}. Conformal risk control
extends these guarantees to any monotone loss~\cite{angelopoulos2024crc}, and Learn-then-Test certifies general risk constraints through
multiple testing~\cite{angelopoulos2025ltt}. In medical imaging, however, conformal predictors have proved unreliable under shifts in the
input and label distributions~\cite{mehrtens2023pitfalls}, which motivates combining risk control with explicit checks on the
acquisition conditions of each image.

Agentic clinical systems built on large language models plan and call tools: MDAgents adapts the collaboration structure of several
models to case complexity~\cite{kim2024mdagents}, and MedRAX orchestrates chest radiograph analysis tools~\cite{fallahpour2025medrax}.
Their tool use is driven by language rather than by cost, and their decisions carry no certified error rate. Explanations, finally,
are often presented as evidence for trust, yet saliency methods such as Grad-CAM~\cite{selvaraju2017gradcam} and Integrated
Gradients~\cite{sundararajan2017axiomatic} localise pathology poorly in chest radiograph benchmarks~\cite{saporta2022benchmarking},
and attention rollout~\cite{abnar2020quantifying} was introduced to quantify information flow in transformers rather than to
localise pathology. The present work treats verification, image quality and explanations as candidate evidence
for a triage decision and measures, under matched budgets and equal certified safety, whether each improves on uncertainty alone.

\section{Materials and Methods}\label{sec:method}

\subsection{Dataset Description}\label{sec:data}
\begin{figure}[t]
\centering
\includegraphics[width=\linewidth]{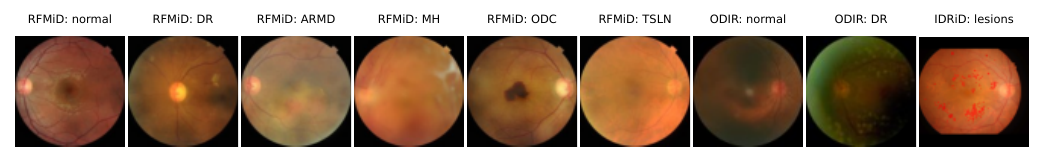}
\caption{Example images after field-of-view cropping: RFMiD test images with single labels, ODIR-5K images, and an IDRiD image with
its lesion masks overlaid.}
\label{fig:samples}
\end{figure}

Three public datasets are used, each for a single, predefined purpose (Table~\ref{tab:datasets}; examples in Fig.~\ref{fig:samples}).
RFMiD~\cite{pachade2021rfmid} contains 3200 colour fundus photographs acquired with three fundus cameras and annotated by two
senior retinal experts, released as official training
(1920), validation (640) and test (640) splits with \nLabOrig{} binary labels including a screening label (normal or abnormal).
Labels with fewer than \minPos{} positive training images are excluded, a rule computed on the training split only, which leaves
\nLabRet{} labels. An iteratively stratified 20\% subset of the official training split is held out as a calibration
set; it is never used for training, early stopping or fitting of any model and serves only to certify the acceptance threshold.
ODIR-5K~\cite{li2021odir} provides per-eye diagnostic keywords from a different population and set of cameras. The screening label is
derived from these keywords, and diseases are mapped to RFMiD labels only where the correspondence is unambiguous (\odirMapped);
glaucoma and cataract are not mapped. ODIR-5K is used for external evaluation without retraining. The segmentation subset of
IDRiD~\cite{porwal2018idrid} provides pixel-level masks of microaneurysms, haemorrhages, hard and soft exudates and the optic disc,
and is used only to validate explanations. A leakage audit based on file hashes and high-pass thumbnail correlation, including
horizontally flipped copies, found \leakExact{} exact and no near-duplicate images across RFMiD partitions.

\begin{table}[t]
\caption{Datasets, partitions and their use. Abnormal prevalence: 79.1\% in the RFMiD test split. Lesion masks: microaneurysms, haemorrhages, hard and soft exudates, optic disc.}
\label{tab:datasets}\centering\scriptsize
\begin{tabular*}{\textwidth}{@{\extracolsep{\fill}}llrll@{}}
\toprule
Dataset & Partition & Images & Use & Labels \\
\midrule
RFMiD~\cite{pachade2021rfmid} & official training, 80\% & 1536 & model training & screening + 24 diseases \\
 & official training, 20\% & 384 & conformal calibration only & same \\
 & official validation & 640 & model selection and calibration & same \\
 & official test & 640 & evaluation only & same \\
ODIR-5K~\cite{li2021odir} & labelled eyes & 6690 & external evaluation only & screening + 12 diseases \\
IDRiD~\cite{porwal2018idrid} & segmentation set & 81 & explanation validation only & lesion masks \\
\bottomrule
\end{tabular*}
\end{table}

\subsection{Problem Formulation}\label{sec:problem}
Let $x$ be a fundus photograph with label vector $y\in\{0,1\}^{L}$, where the first component is the screening label and the remaining
$L-1$ components are diseases. A primary model produces calibrated probabilities $\hat p(x)$ and decisions
$\hat y_\ell=\mathbb 1[\hat p_\ell\ge t_\ell]$ with per-label thresholds $t_\ell$ fitted on validation data. The automated decision is
called unsafe when the screening decision is wrong or at least one present disease is missed,
\begin{equation}
E(x)=\mathbb 1[\hat y_1\neq y_1]\;\lor\;\mathbb 1[\exists \ell>1: y_\ell=1,\ \hat y_\ell=0].
\end{equation}
False-positive findings are not counted as unsafe because they lead to review. Each case is accepted, verified (accepted after
additional evidence) or referred, and the system is evaluated by its unsafe-acceptance and referral rates.

\subsection{Agent Design and Evidence}\label{sec:agents}
\begin{sidewaysfigure}
\centering
\includegraphics[width=\textheight]{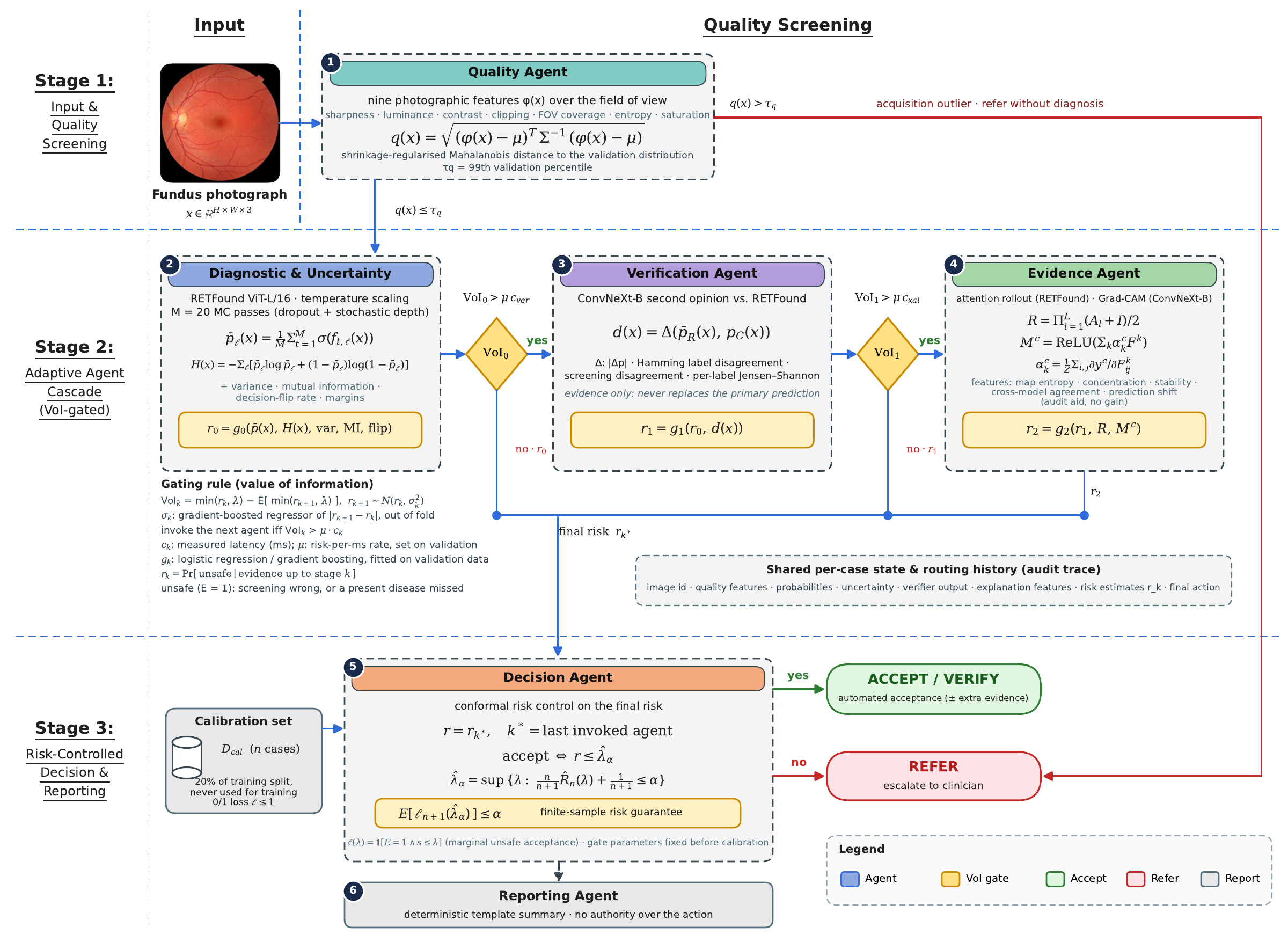}
\caption{RetinaGuard agent graph. The quality agent refers acquisition outliers before diagnosis; the diagnostic, verification and
evidence agents run in sequence, and after each stage a failure predictor estimates the probability $r_k(x)$ that the automated
decision is unsafe. The next agent is invoked only when its expected value of information exceeds its cost $\mu c$. The decision agent
accepts a case only when its final risk is below the threshold $\hat\lambda_\alpha$ certified by conformal risk control, and a reporting
agent summarises the evidence without authority over the action.}
\label{fig:arch}
\end{sidewaysfigure}

RetinaGuard is organised as a graph of agents that share a structured per-case state (Fig.~\ref{fig:arch}, Table~\ref{tab:agents}); an
agent is a component that decides where a case goes next or whether to spend computation on it.

The quality agent computes nine photographic features over the field of view and summarises them by a shrinkage-regularised
Mahalanobis distance $q(x)$ to the validation distribution; images above the 99th validation percentile are referred as acquisition
outliers before any diagnostic model is run. The diagnostic agent applies RETFound~\cite{zhou2023retfound}, fine-tuned for multi-label
RFMiD classification with sigmoid outputs, followed by global or per-label temperature scaling~\cite{guo2017calibration} selected by
cross-validated likelihood. The uncertainty agent runs 20 stochastic forward passes with dropout~\cite{gal2016dropout} and stochastic
depth active and records predictive variance, mutual information and the fraction of passes whose label set differs from the
deterministic decision, from which a stage-0 failure predictor estimates $r_0(x)=\Pr[E=1\mid\text{stage-0 evidence}]$.

The verification agent applies an independently trained ConvNeXt-B~\cite{liu2022convnext}. Its prediction never replaces the primary
prediction; absolute probability differences, label and screening disagreement and per-label Jensen--Shannon divergence feed a stage-1
predictor $r_1(x)$. The evidence agent computes attention rollout~\cite{abnar2020quantifying} for RETFound, adapted to its average-pooled
head, the same map under a mild label-preserving perturbation, and Grad-CAM~\cite{selvaraju2017gradcam} for the verifier; map entropy,
concentration, stability, cross-model agreement and the prediction shift under perturbation feed a stage-2 predictor $r_2(x)$. Each
failure predictor is a logistic regression or gradient-boosted tree, chosen by five-fold cross-validated AUROC on validation data. The
decision agent refers a case when its final risk exceeds a certified threshold (Sect.~\ref{sec:crc}), and a reporting agent renders a
deterministic summary with no influence on the action.

\begin{table}[t]
\caption{Agents, tools and evidence. Latency is the measured per-image cost of each tool on one T4 GPU (batch size 1).}
\label{tab:agents}\centering\scriptsize
\begin{tabularx}{\textwidth}{@{}l>{\raggedright\arraybackslash}p{2.4cm}Yr@{}}
\toprule
Agent & Tool & Evidence passed on & ms \\
\midrule
Quality & photographic statistics & sharpness, luminance, contrast, clipping, field-of-view coverage, entropy, saturation; outlier distance $q(x)$ & \msQual \\
Diagnostic & RETFound ViT-L/16 & calibrated probabilities, margins, entropy, predicted label count & \msDet \\
Uncertainty & 20 Monte Carlo passes & variance, mutual information, decision-flip rate; risk $r_0$ & \msMc \\
Verification & ConvNeXt-B & probability difference, label disagreement, Jensen--Shannon divergence; risk $r_1$ & \msVer \\
Evidence & rollout, Grad-CAM & map entropy and concentration, stability, cross-model agreement; risk $r_2$ & \msXai \\
Decision & certified threshold & accept, verify or refer & -- \\
Reporting & template & structured summary (no authority over the action) & -- \\
\bottomrule
\end{tabularx}
\end{table}

\subsection{Cost-Aware Tool Invocation}\label{sec:voi}
The verifier and the evidence agent are invoked only when their evidence is expected to change the decision enough to justify
their cost. With the decision loss $\ell(\text{accept})=r$ and
$\ell(\text{refer})=\lambda$, where $\lambda$ is the referral cost expressed on the risk scale, the myopic expected value of
information of a tool that updates $r_k$ to $r_{k+1}$ is
\begin{equation}
\mathrm{VoI}_k(x)=\min(r_k,\lambda)-\mathbb E\big[\min(r_{k+1},\lambda)\big].
\end{equation}
Modelling $r_{k+1}\sim\mathcal N(r_k,\sigma_k^2(x))$, where $\sigma_k(x)$ is predicted from stage-$k$ evidence by a gradient-boosted
regressor of $|r_{k+1}-r_k|$ fitted out of fold on validation data, gives the closed form
$\mathrm{VoI}_k=\min(r_k,\lambda)-r_k+\sigma_k\varphi(d)+(r_k-\lambda)\Phi(d)$ with $d=(r_k-\lambda)/\sigma_k$. A tool is invoked when
$\mathrm{VoI}_k(x)>\mu c_k$, where $c_k$ is its measured latency and $\mu$ is a single exchange rate between risk and milliseconds.
The rate $\mu$ is chosen on validation data as the cheapest setting whose mean error-detection rate over referral budgets of 10--40\%
remains within one percentage point of invoking every tool for every case.

\subsection{Risk-Controlled Acceptance Threshold}\label{sec:crc}
The acceptance threshold is calibrated on the held-out calibration set. Because the per-case loss $\ell_i(\lambda)=\mathbb 1[E_i=1\wedge s_i\le\lambda]$, an unsafe automated acceptance, is monotone in
$\lambda$, conformal risk control~\cite{angelopoulos2024crc} selects
$\hat\lambda=\sup\{\lambda: \tfrac{n}{n+1}\hat R_n(\lambda)+\tfrac{1}{n+1}\le\alpha\}$ and guarantees
$\mathbb E[\ell_{n+1}(\hat\lambda)]\le\alpha$ for exchangeable test cases~\cite{angelopoulos2023gentle}. Because the
gate parameters are fixed before calibration, the guarantee holds for the adaptive policy regardless of which tools a case triggered.
A Learn-then-Test procedure~\cite{angelopoulos2025ltt} with fixed-sequence testing additionally certifies the selective risk with
probability $1-\delta$. Applying the same certification to every baseline compares methods by their coverage at equal certified safety.

\section{Experimental Results}\label{sec:results}

\subsection{Experimental Setup}\label{sec:setup}
All backbones, namely ResNet-50~\cite{he2016resnet}, EfficientNet-B3~\cite{tan2019efficientnet}, ConvNeXt-B~\cite{liu2022convnext},
Swin-B~\cite{liu2021swin} and RETFound ViT-L/16~\cite{zhou2023retfound}, receive the same field-of-view-cropped $224\times224$ inputs
and the same mild augmentation (rotation, scaling, horizontal flipping, brightness and contrast). Training uses class-weighted binary
cross-entropy, AdamW, a cosine schedule with warm-up, mixed precision and early stopping on validation macro-AUROC. RETFound is
fine-tuned end to end following its original recipe (40 epochs, five warm-up epochs, base learning rate $5\times10^{-3}$, layer-wise
decay 0.65); a conservative staged schedule with frozen early blocks yields a weaker primary model, with which the complete analysis is
repeated (Sect.~\ref{sec:strength}). Two further RETFound seeds estimate seed variance. Training and all GPU inference were executed on
Kaggle with two NVIDIA Tesla T4 GPUs (PyTorch 2.11, timm 1.0). Every tool was run once for every image and its outputs stored; the
policy, risk models, conformal calibration and statistics were then computed on a CPU, so routing decisions are replayed exactly from
measured outputs, and computational cost is the measured per-call latency of each tool times the number of policy invocations.

\subsection{Baselines and Evaluation Metrics}\label{sec:baselines}
All referral methods are compared at identical referral budgets $b$ of 5, 10, 15, 20, 30 and 40\%: each method refers exactly
the $\lfloor bn\rceil$ test cases with the highest scores. Table~\ref{tab:baselines} lists the methods; B1 and B2 select their score on
validation data among the candidates shown. Table~\ref{tab:metrics} defines the metrics. Paired differences between methods are
assessed with \nBoot{} case-level bootstrap resamples, and exact McNemar tests on the captured errors; Holm correction across budgets is applied to the bootstrap p-values.

\begin{table}[t]
\caption{Referral methods compared at matched budgets.}\label{tab:baselines}\centering\scriptsize
\begin{tabularx}{\textwidth}{@{}lY@{}}
\toprule
Method & Referral score \\
\midrule
B1 & lowest confidence (raw screening confidence or minimum label margin) \\
B2 & calibrated uncertainty (entropy, label margin, Monte Carlo mutual information, variance or flip rate) \\
B3 & unlearned sum of standardised uncertainty and verifier disagreement \\
B4 & learned failure predictor without explanation features, verifier invoked for every case \\
B5 & RetinaGuard: learned failure predictor with value-of-information tool invocation \\
B6 & learned failure predictor with all evidence computed for every case \\
\bottomrule
\end{tabularx}
\end{table}

\begin{table}[t]
\caption{Evaluation metrics. $A$: accepted (not referred) cases; $R$: referred cases; $E_i$: unsafe automated decision; $n$: test cases.}
\label{tab:metrics}\centering\scriptsize
\begin{tabularx}{\textwidth}{@{}>{\raggedright\arraybackslash}p{3.3cm}>{\raggedright\arraybackslash}p{3.6cm}Y@{}}
\toprule
Metric & Definition & Interpretation \\
\midrule
Coverage & $|A|/n$ & share of cases decided automatically \\
Unsafe acceptance (selective risk) & $\sum_{i\in A}E_i/|A|$ & error rate among accepted cases \\
Marginal unsafe rate & $\sum_{i\in A}E_i/n$ & quantity bounded by conformal risk control \\
Error-detection rate & $\sum_{i\in R}E_i/\sum_{i}E_i$ & share of unsafe decisions that are referred \\
Error-detection AUROC & AUROC of the score for $E_i$ & ranking quality of a failure predictor \\
Macro-AUROC / macro-F1 & mean over labels & classification quality (all or accepted cases) \\
ECE & $\sum_b \tfrac{|B_b|}{n}\,|\bar y_b-\bar p_b|$ & calibration error over 15 probability bins \\
\bottomrule
\end{tabularx}
\end{table}

\subsection{Backbone Performance and Calibration}
Table~\ref{tab:backbones} compares the backbones on the official test split. RETFound reaches a macro-AUROC of \aucRf{} (95\% CI
\aucRfLo--\aucRfHi; three-seed mean \aucRfSeedMean$\pm$\aucRfSeedSd) and a screening AUROC of \scrRf. Its original fine-tuning recipe
outperforms the conservative staged schedule (\aucRfCons), a gain of similar size to the gap between
backbones. ConvNeXt-B (\aucCnx) and Swin-B (\aucSwin), both pretrained on natural images, are significantly stronger on macro-AUROC
(paired bootstrap, $p<0.01$), consistent with a recent comparison in which a natural-image foundation model outperformed RETFound for
ocular disease detection~\cite{hou2026dinov2}. RETFound is retained as the primary model because the study concerns when a given model
may act alone rather than which model is most accurate; the architecturally different ConvNeXt-B serves as the verifier. Temperature
scaling, selected by cross-validated likelihood, leaves the screening ECE essentially unchanged (\eceRaw{} before and \eceCal{} after).
Under the definition of Sect.~\ref{sec:problem}, \errTest\% of automated test decisions are unsafe (\nErrTest{} of \nTest), so the triage
problem is substantial.

\begin{table}[t]
\caption{Backbones on the RFMiD test split ($n=\nTest$). Macro-AUROC over \nLabRet{} labels with 95\% bootstrap confidence interval.}
\label{tab:backbones}\centering\scriptsize
\begin{tabular}{lccc}
\toprule
Backbone & Macro-AUROC [95\% CI] & Screening AUROC & Macro-F1 \\
\midrule
ResNet-50 & \aucRes{} [\aucResLo, \aucResHi] & \scrRes & \foneRes \\
EfficientNet-B3 & \aucEff{} [\aucEffLo, \aucEffHi] & \scrEff & \foneEff \\
ConvNeXt-B (verifier) & \aucCnx{} [\aucCnxLo, \aucCnxHi] & \scrCnx & \foneCnx \\
Swin-B & \aucSwin{} [\aucSwinLo, \aucSwinHi] & \scrSwin & \foneSwin \\
RETFound, conservative schedule & \aucRfCons{} [\aucRfConsLo, \aucRfConsHi] & \scrRfCons & \foneRfCons \\
RETFound, original recipe (primary) & \aucRf{} [\aucRfLo, \aucRfHi] & \scrRf & \foneRf \\
\bottomrule
\end{tabular}
\end{table}

\subsection{Failure Prediction and Evidence Ablation}
Table~\ref{tab:ablation} adds evidence cumulatively. The minimum label margin detects failures moderately (error-detection AUROC
\aucAz); calibrated entropy is slightly better (\errAucEnt) and Monte Carlo mutual information worse (\errAucMI). A learned combination
of uncertainty features reaches \aucAone, a significant gain ($\Delta=\dUnc$, 95\% CI \dUncLo{} to \dUncHi). Verifier disagreement
predicts unsafe decisions on its own (AUROC \verAuc), and RETFound is unsafe while ConvNeXt-B is not in \vcRwVc{} test cases, yet it adds
almost nothing to the uncertainty features ($\Delta=\dVerif$, CI \dVerifLo{} to \dVerifHi): most of what the verifier reveals is
already reflected in the uncertainty of the primary model. Quality ($\Delta=\dQual$) and explanation features ($\Delta=\dXai$) add no
significant information on in-distribution images either, so the gain over confidence is attributable to the uncertainty group.

\begin{table}[t]
\caption{Cumulative evidence ablation on the test split; failure predictors are fitted on validation data only. EDR@20: share of the
\nErrTest{} unsafe decisions referred at a 20\% referral budget.}\label{tab:ablation}
\centering\scriptsize
\begin{tabular}{lcc}
\toprule
Evidence used by the failure predictor & Error-detection AUROC & EDR@20 (\%) \\
\midrule
A0 minimum label margin (no learning) & \aucAz & \edrAz \\
A1 uncertainty (calibrated and Monte Carlo) & \aucAone & \edrAone \\
A2 A1 + verifier disagreement & \aucAtwo & \edrAtwo \\
A3 A2 + image quality & \aucAthree & \edrAthree \\
A4 A3 + explanation features & \aucAfour & \edrAfour \\
\bottomrule
\end{tabular}
\end{table}

\subsection{Comparison at Matched Referral Budgets}
\begin{figure}[t]
\centering
\includegraphics[width=\linewidth]{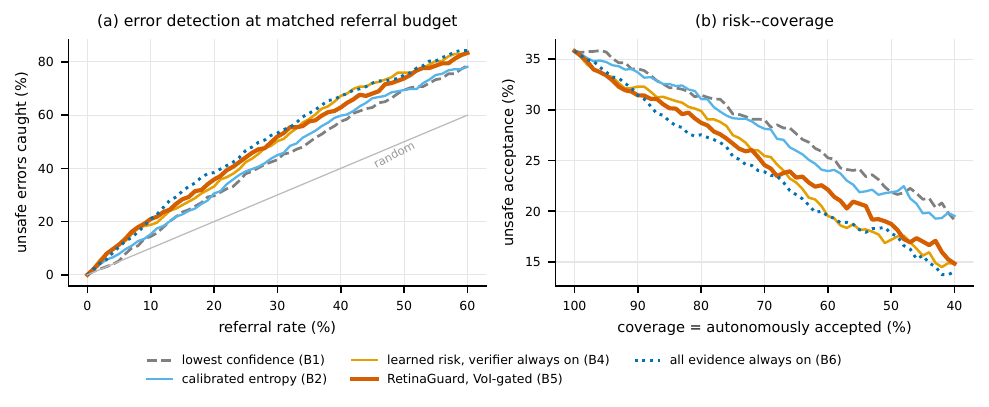}
\caption{Comparison at matched referral budgets on the RFMiD test split. (a) Share of unsafe automated decisions captured by referral;
(b) unsafe-acceptance rate among the cases left to the system.}
\label{fig:selective}
\end{figure}

Fig.~\ref{fig:selective} and Table~\ref{tab:matched} compare methods that refer exactly the same number of cases. At a 10\% budget,
lowest-confidence referral (B1) captures \capBoneTen{} of the \nErrTest{} unsafe decisions and RetinaGuard (B5) \capBfiveTen{}
($\Delta\text{EDR}=\dEdrTen$ percentage points, CI \dEdrTenLo{} to \dEdrTenHi), lowering unsafe acceptance from \srBoneTen\% to
\srBfiveTen\%. RetinaGuard captures more errors than B1 and calibrated uncertainty (B2) at every tested budget, and the bootstrap interval
against B1 excludes zero at \nCiBudgets{} of \nBudgets{} budgets, but after Holm correction the difference is significant only against B1
at the 5\% budget (\capBfiveFive{} versus \capBoneFive{} errors, $p\pEdrFive$; unadjusted exact McNemar $p\pMcFive$) and at no budget
against B2 (smallest adjusted $p\minHolmTwo$). RetinaGuard is statistically indistinguishable from the learned predictor with the
verifier always on (B4) and from computing all evidence for every case (B6) at every budget. With this primary model the improvement
over single-score referral is consistent in direction but modest, and it stems from the learned failure predictor rather than from the
additional agents.

\begin{table}[t]
\caption{Errors captured (out of \nErrTest) and unsafe acceptance among accepted cases at matched referral budgets; without referral the
unsafe-acceptance rate is \srBase\%.}\label{tab:matched}
\centering\scriptsize
\setlength{\tabcolsep}{5pt}\begin{tabular}{@{}lcccccc@{}}
\toprule
& \multicolumn{2}{c}{Budget 10\%} & \multicolumn{2}{c}{Budget 20\%} & \multicolumn{2}{c}{Budget 30\%} \\
\cmidrule(lr){2-3}\cmidrule(lr){4-5}\cmidrule(lr){6-7}
Method & Caught & Unsafe (\%) & Caught & Unsafe (\%) & Caught & Unsafe (\%) \\
\midrule
B1 lowest confidence & \capBoneTen & \srBoneTen & \capBoneTwenty & \srBoneTwenty & \capBoneThirty & \srBoneThirty \\
B2 calibrated uncertainty & \capBtwoTen & \srBtwoTen & \capBtwoTwenty & \srBtwoTwenty & \capBtwoThirty & \srBtwoThirty \\
B4 verifier always on & \capBfourTen & \srBfourTen & \capBfourTwenty & \srBfourTwenty & \capBfourThirty & \srBfourThirty \\
B6 all evidence always on & \capBsixTen & \srBsixTen & \capBsixTwenty & \srBsixTwenty & \capBsixThirty & \srBsixThirty \\
B5 RetinaGuard & \capBfiveTen & \srBfiveTen & \capBfiveTwenty & \srBfiveTwenty & \capBfiveThirty & \srBfiveThirty \\
\bottomrule
\end{tabular}
\end{table}

\subsection{Certified Operating Points}
Table~\ref{tab:crc} reports the operating points obtained when each method's threshold is calibrated on the held-out calibration
set. At $\alpha=5\%$ RetinaGuard accepts \covBfiveFive\% of test cases (realised marginal unsafe rate \muBfiveFive\%), compared with
\covBoneFive\% for lowest-confidence referral and \covBtwoFive\% for calibrated uncertainty, whose realised rate of \muBtwoFive\% exceeds
the target. At $\alpha=10\%$ calibrated uncertainty accepts the most cases, but its realised rate (\muBtwoTen\%) and that of
lowest-confidence referral (\muBoneTen\%) exceed the nominal level, whereas RetinaGuard stays below it (\muBfiveTen\%). Such single-split
deviations are expected because the guarantee holds in expectation; over random re-splits the realised rate of RetinaGuard averaged
\crcMean\% at $\alpha=5\%$ (90th percentile \crcQnine\%). Learn-then-Test certifies a selective risk below \lttAlpha\% with probability
0.9 at \lttCov\% coverage.

\begin{table}[t]
\caption{Operating points certified by conformal risk control on the calibration set: coverage and realised marginal unsafe rate
on the test split (both in \%).}\label{tab:crc}
\centering\scriptsize
\setlength{\tabcolsep}{5pt}\begin{tabular}{@{}lcccccc@{}}
\toprule
& \multicolumn{2}{c}{$\alpha=2\%$} & \multicolumn{2}{c}{$\alpha=5\%$} & \multicolumn{2}{c}{$\alpha=10\%$} \\
\cmidrule(lr){2-3}\cmidrule(lr){4-5}\cmidrule(lr){6-7}
Method & Coverage & Unsafe & Coverage & Unsafe & Coverage & Unsafe \\
\midrule
B1 lowest confidence & \covBoneTwo & \muBoneTwo & \covBoneFive & \muBoneFive & \covBoneTen & \muBoneTen \\
B2 calibrated uncertainty & \covBtwoTwo & \muBtwoTwo & \covBtwoFive & \muBtwoFive & \covBtwoTen & \muBtwoTen \\
B4 verifier always on & \covBfourTwo & \muBfourTwo & \covBfourFive & \muBfourFive & \covBfourTen & \muBfourTen \\
B5 RetinaGuard & \covBfiveTwo & \muBfiveTwo & \covBfiveFive & \muBfiveFive & \covBfiveTen & \muBfiveTen \\
\bottomrule
\end{tabular}
\end{table}

\subsection{Dependence on Primary-Model Strength}\label{sec:strength}
The full analysis was repeated with the RETFound model fine-tuned with the conservative schedule, which is weaker than the verifier
by a wider margin (Table~\ref{tab:strength}). With this primary model, verifier disagreement adds significant information to the
uncertainty features ($\Delta\text{AUROC}=\cmpWdVer$, CI \cmpWdVerLo{} to \cmpWdVerHi, $p\cmpWpVer$), RetinaGuard captures
significantly more errors than lowest-confidence referral at \cmpWnSigOne{} of \cmpWnB{} budgets and than calibrated uncertainty at
\cmpWnSigTwo{} of \cmpWnB{} after Holm correction, and at the nominal $\alpha=5\%$ operating point it accepts \cmpWcovBfive\% of cases
(realised unsafe rate \cmpWmuBfive\%) compared with
\cmpWcovBone\% for lowest-confidence referral and \cmpWcovBtwo\% for calibrated uncertainty. With the stronger primary model the same
comparisons give \cmpSnSigOne{} and \cmpSnSigTwo{} significant budgets and nearly equal certified coverage. Two quantities remain stable
across both settings: the nominal coverage of RetinaGuard at $\alpha=5\%$ (\cmpWcovBfive\% and \cmpScovBfive\%) and its computational
saving (\cmpWsave\% and \cmpSsave\%). The quality agent lowered the worst-case unsafe rate under corruption in both settings, from
\cmpWmuNoQ\% to \cmpWmuMax\% and from \cmpSmuNoQ\% to \cmpSmuMax\%, but only for the weaker model did it stay near the target. The value-of-information gate adapts without retuning: it calls the verifier for \cmpWver\% of test cases
when disagreement is informative and for \cmpSver\% when it is not. The benefit of the agentic layer over single-score referral
therefore depends on how much the primary model's own uncertainty already reveals about its errors.

\begin{table}[t]
\caption{The same analysis with a weaker and a stronger RETFound primary model. EDR gain: difference in error-detection rate between
RetinaGuard (B5) and lowest-confidence (B1) or calibrated-uncertainty (B2) referral at a 20\% budget, in percentage points, with
Holm-adjusted p-value. Coverage: share of test cases accepted at the certified $\alpha=5\%$ level.}\label{tab:strength}
\centering\scriptsize
\begin{tabularx}{\textwidth}{@{}Ycc@{}}
\toprule
Quantity & Conservative schedule & Original recipe \\
\midrule
Primary macro-AUROC & \cmpWauc & \cmpSauc \\
Unsafe automated decisions (\%) & \cmpWerr{} (\cmpWnErr) & \cmpSerr{} (\cmpSnErr) \\
Error-detection AUROC, A0 / A1 / A2 & \cmpWAz{} / \cmpWAone{} / \cmpWAtwo & \cmpSAz{} / \cmpSAone{} / \cmpSAtwo \\
Gain from verifier disagreement [95\% CI] & \cmpWdVer{} [\cmpWdVerLo, \cmpWdVerHi] & \cmpSdVer{} [\cmpSdVerLo, \cmpSdVerHi] \\
EDR gain over B1 at 20\% ($p$) & \cmpWdOneTwenty{} (\cmpWpOneTwenty) & \cmpSdOneTwenty{} (\cmpSpOneTwenty) \\
EDR gain over B2 at 20\% ($p$) & \cmpWdTwoTwenty{} (\cmpWpTwoTwenty) & \cmpSdTwoTwenty{} (\cmpSpTwoTwenty) \\
Significant budgets vs B1 / vs B2 & \cmpWnSigOne{} / \cmpWnSigTwo{} of \cmpWnB & \cmpSnSigOne{} / \cmpSnSigTwo{} of \cmpSnB \\
Coverage at $\alpha=5\%$: B1 / B2 / B5 (\%) & \cmpWcovBone{} / \cmpWcovBtwo{} / \cmpWcovBfive & \cmpScovBone{} / \cmpScovBtwo{} / \cmpScovBfive \\
Verifier calls by RetinaGuard (\%) & \cmpWver & \cmpSver \\
Worst unsafe rate under corruption, no gate / gate (\%) & \cmpWmuNoQ{} / \cmpWmuMax & \cmpSmuNoQ{} / \cmpSmuMax \\
\bottomrule
\end{tabularx}
\end{table}

\subsection{Robustness and the Quality Agent}\label{sec:robust}
\begin{figure}[t]
\centering
\includegraphics[width=\linewidth]{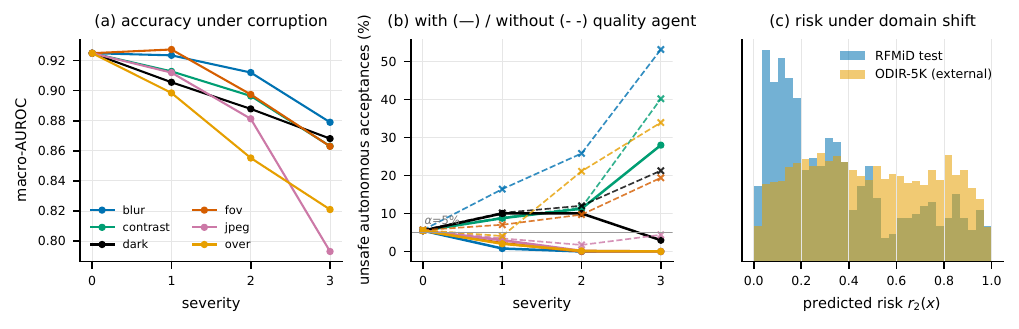}
\caption{(a) Macro-AUROC under six corruptions (severity 0: clean). (b) Marginal unsafe rate at the $\alpha=5\%$ operating point with
(solid) and without (dashed) the quality agent. (c) Distribution of predicted risk on the RFMiD test split and on ODIR-5K.}
\label{fig:robust}
\end{figure}

Macro-AUROC declines gradually under corruption (Fig.~\ref{fig:robust}a), but the unsafe-error rate rises sharply (\errblurThree\%
under severe blur). The learned risk rises under darkening and loss of field of view but falls at the most severe levels of blur,
contrast reduction, compression and overexposure (severe blur: \riskblurThree{} against \riskClean{} on clean images), and Monte Carlo
mutual information changes little. Without the quality agent the marginal unsafe rate at the $\alpha=5\%$ operating point reaches
\muNoQmax\%; with it, strongly degraded images are referred before diagnosis and the rate stays at or below \muMax\% (Fig.~\ref{fig:robust}b).
The worst remaining case is severe contrast reduction (\mucontrastThree\%), which the photographic features detect only in part.

\subsection{External Validation on ODIR-5K}
Without retraining, the screening AUROC falls from \scrAucR{} on RFMiD to \scrAucO{} on ODIR-5K, partly because the keyword-derived
labels are noisy. The learned risk shifts upwards (mean \riskR{} to \riskO; Cliff's $\delta=\cdRisk$, $p\pShiftRisk$;
Fig.~\ref{fig:robust}c), whereas Monte Carlo mutual information does not (\miR{} versus \miO). The policy refers \refO\% of ODIR-5K
images, yet the realised marginal unsafe rate of \muO\% exceeds the 5\% target, and the risk ranks ODIR-5K errors only weakly (AUROC
\riskAucO{} versus \riskAucR{} in distribution): the guarantee does not transfer to the new population.

\subsection{Explainability Validation}
\begin{figure}[t]
\centering
\includegraphics[width=0.92\linewidth]{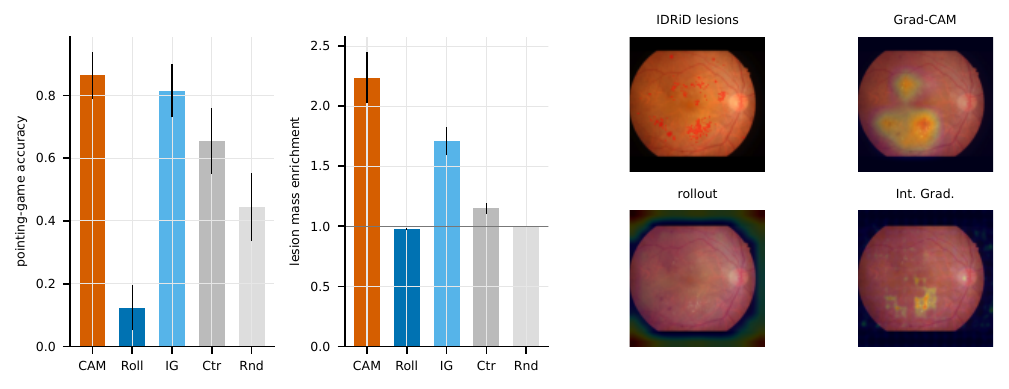}
\caption{Explanation validation on IDRiD lesion masks (union of microaneurysms, haemorrhages and exudates). CAM: Grad-CAM (ConvNeXt-B);
Roll: attention rollout and IG: Integrated Gradients (RETFound); Ctr: centre prior; Rnd: random map. Bars show means with 95\%
confidence intervals.}
\label{fig:xai}
\end{figure}

On \nIdrid{} IDRiD images with lesion masks (Fig.~\ref{fig:xai}), Grad-CAM of the ConvNeXt-B verifier localises lesions well
(pointing-game accuracy \ptCam; lesion attribution mass \enrCam{} times that of a uniform map), and Integrated
Gradients~\cite{sundararajan2017axiomatic} of RETFound also exceeds both a random map and a centre prior (pointing \ptIg; \enrIg{} times
uniform). Attention rollout of RETFound does not localise lesions (\ptRoll; \enrRoll{} times uniform, below both the random map and the
centre prior), although its maps are highly stable under perturbation (Spearman \stabRoll), which shows that stability does not imply
faithfulness. Gradient-based attributions of the same transformer are therefore informative where attention rollout is not, which
argues against presenting attention maps as clinical evidence.

\section{Discussion}\label{sec:discussion}
A verifier helps when it is right where the primary model is wrong and the primary model does not know it. With the conservatively
fine-tuned RETFound, which trailed ConvNeXt-B by a wide margin, disagreement exposed errors that uncertainty missed, and RetinaGuard
outperformed single-score referral at most budgets and at the nominal operating point. With the properly fine-tuned model, both models
were wrong on \vcBoth{} of the \nErrTest{} unsafe cases, uncertainty already carried most of the signal, and the gain shrank to a few
percentage points. The coverage of RetinaGuard barely changed; the baselines became stronger. Quality and explanation features added no
significant information in either setting.

Three components proved useful regardless of model strength. The value-of-information gate kept the error detection of always-on
verification while calling the verifier for \cmpWver\% or \cmpSver\% of cases and never calling the explanation agent; executing every
tool costs \msFixed{} ms per image against \msAgent{} ms for the gated policy, and Monte Carlo uncertainty estimation accounts for most of
the remaining cost. The quality agent prevented unsafe acceptances on severely blurred, overexposed, darkened or cropped photographs that
model-internal signals did not prevent, although severe contrast loss was only partly detected. Conformal risk control produced
operating points close to their nominal levels in distribution (\muBfiveFive\% on the test split and \crcMean\% on average over
re-splits at $\alpha=5\%$), but not on ODIR-5K, where model uncertainty did not register the shift and the realised unsafe rate exceeded
the target despite referral of most images.

The study has limitations. All development data come from one public dataset with \nTest{} test images, which limits the power to detect
differences of a few percentage points, and the external labels are derived from diagnostic keywords. The two primary models differ only
in their fine-tuning schedule, so the dependence on model strength rests on two points rather than a systematic sweep. The quality agent
was validated with synthetic corruptions rather than graded gradability labels. The conformal guarantee is marginal, assumes
exchangeability and does not hold under population shift. The definition of an unsafe decision weights all missed labels equally,
although cost-weighted losses fit directly into the same framework. Finally, computational cost is derived from measured per-tool
latencies rather than from an end-to-end deployment.

\section{Conclusion}\label{sec:conclusion}
This paper presented RetinaGuard, an agentic triage framework that decides for each fundus photograph whether an automated
multi-disease decision may stand, which additional evidence is worth acquiring and when the case must be referred, with the acceptance
threshold certified by conformal risk control. Evaluated on RFMiD at matched referral budgets and equal certified safety with two RETFound
primary models, the framework improved substantially on confidence and calibrated-uncertainty referral when the primary model was weaker
than its verifier, because verifier disagreement exposed errors that uncertainty missed, and offered only a small advantage once the
primary model was properly fine-tuned and its uncertainty already explained most of its errors. In both settings it maintained certified
coverage near \cmpScovBfive\%, reduced computation by about 30\% by calling the verifier only when the expected value of its evidence
justified the cost, and, through its quality agent, substantially reduced unsafe acceptances under most image corruptions, although severe contrast loss was
only partly detected, while guarantees did not transfer to the external ODIR-5K population. Future work will measure the dependence on model
strength systematically across primary models and verifiers, validate the quality agent against clinician-graded gradability, introduce
clinically weighted losses and conformal methods that adapt to distribution shift, and evaluate the framework on larger multi-site
cohorts.

\bibliographystyle{splncs04-unsrt}
\bibliography{refs}
\end{document}